\documentclass[11pt,a4paper]{article}
\usepackage[hyperref]{eacl2021}
\usepackage{times}
\usepackage{latexsym}

\usepackage{microtype}
\usepackage{graphicx}
\usepackage{hyperref}

\aclfinalcopy 

\title{MadDog:\\A Web-based System for Acronym Identification and Disambiguation}

\author{Amir Pouran Ben Veyseh\textsuperscript{\rm 1}, Franck Dernoncourt\textsuperscript{\rm 2}, \\
{\bf Walter Chang}\textsuperscript{\rm 2}, {\bf and Thien Huu Nguyen}\textsuperscript{\rm 1} \\
\textsuperscript{\rm 1} Department of Computer and Information Science, University of Oregon,
\\Eugene, OR 97403, USA\\
\textsuperscript{\rm 2} Adobe Research, San Jose, CA, USA\\
  \texttt{\{apouranb,thien\}@cs.uoregon.edu}, \\ \texttt{\{franck.dernoncourt,wachang\}@adobe.com}
}

\date{}

\begin{document}
\maketitle
\begin{abstract}
Acronyms and abbreviations are the short-form of longer phrases and they are ubiquitously employed in various types of writing. Despite their usefulness to save space in writing and reader's time in reading, they also provide challenges for understanding the text especially if the acronym is not defined in the text or if it is used far from its definition in long texts. To alleviate this issue, there are considerable efforts both from the research community and software developers to build systems for identifying acronyms and finding their correct meanings in the text. However, none of the existing works provide a unified solution capable of processing acronyms in various domains and to be publicly available. Thus, we provide the first web-based acronym identification and disambiguation system which can process acronyms from various domains including scientific, biomedical, and general domains. The web-based system is publicly available at \url{http://iq.cs.uoregon.edu:5000} and a demo video is available at
\url{https://youtu.be/IkSh7LqI42M}. The system source code is also available at \url{https://github.com/amirveyseh/MadDog}.
\end{abstract}

\section{Introduction}

Textual contents such as books, articles, reports, and web-blogs in various domains are replete with phrases that are commonly used by people in that field. In order to save space in text writing and also facilitate communication among people who are already familiar with these phrases, the shorthanded form of long phrases, known as acronyms and abbreviations, are frequently used. However, the use of acronyms could also introduce challenges to understand the text, especially for newcomers. More specifically, two types of challenges might hinder reading text with acronyms: 1) In long documents, e.g., a book chapter, an acronym might be defined somewhere in the text and used several times throughout the document. For someone who is not familiar with the definition of the acronym and interested in reading a part of the document, it might be time-consuming to find the definition of the acronym in the document. To solve this problem, an automatic acronym identification tool is required whose goal is to find all acronyms and their definitions that are locally provided in the same document. 2) Some of the acronyms might not be even defined in the document itself. These acronyms are commonly used by writers in a specific domain. To find the correct meaning of them, a reader must look-up the acronym in a dictionary of acronyms. However, due to the shorter length of acronyms compared to their long-form, multiple phrases might be shortened with the same acronym, thereby, they will be ambiguous. In these cases, a deep understanding of the domain is required to recognize the correct meaning of the acronym among all possible long-forms. To solve this issue, a system capable of disambiguating an acronym based on its context is necessary. 

Each of the aforementioned problems, i.e., acronym identification (AI) and acronym disambiguation (AD), has been extensively studied by the research community or software developers. One of the methods which are widely used in acronym identification research is proposed by  \citet{schwartz2002simple}. This is a rule-based model that utilizes character-match between acronym letters and their context to find the acronym and its long-form in text. Later, some feature-based models have been also used for acronym identification \cite{kuo2009bioadi,liu2017multi}. In addition, some of the existing software employs regular expressions for acronym identification in the biomedical domain \cite{BADREX}. Acronym disambiguation is also approached with feature-based models \cite{wang2016clinical} or more advanced deep learning methods \cite{wu2015clinical,ciosici2019unsupervised}. The majority of deep models employ word embeddings to compute the similarity between the candidate long-form and the acronym context. In addition to the existing research for AD, there is some web-based software that employ dictionary look-up to expand an acronym to its long-form \cite{ABBREX}. Note that the methods based on dictionary look-up are not able to disambiguate the acronym if it has multiple meanings. 

Despite the progress made on the AI and AD task in the last two decades, there are some limitations in the prior works that prevent achieving a functional system to be used in practice. More specifically, considering the research on the AD task, all of the prior works employ a small-size dataset covering a few hundred to a few thousand long-forms in a specific domain. Therefore, the models trained in these works are not capable to expand all acronyms of a domain or acronyms in other domains other than the one used in the training set. Although in the recent work \cite{wen2020medal}, authors proposed a big dataset for acronym disambiguation in the medical domain with more than 14 million samples, it is still limited to a specific domain (i.e., medical domain). Another limitation in prior works is that they do not provide a unified system capable of performing both tasks in various domains and to be publicly available. To our knowledge, the only exiting web-based system for AI and AD is proposed by \citet{ciosici2018abbreviation}. For acronym identification, this system employs the rule-based model introduced by \cite{schwartz2002simple}. To handle corner cases, they add extra rules in addition to Schwartz's rules in their system. Unfortunately, they do not provide detailed information about these corner cases and extra rules or any evaluation to assess the performance of the model. For acronym disambiguation, they resort to a statistical model in which a pre-computed vector representation for each candidate long-form is employed to compute the similarity between candidate long-form with the context of the ambiguous acronym represented using another vector. However, there are two limitations with this approach: first, the pre-computed long-form vectors are obtained via only Wikipedia, thus limiting this system to the general domain and incapable of disambiguating acronyms in other domains such as scientific papers or biomedical texts; Second, the AD model based on the pre-computed vectors is a statistical model and is not benefiting from the advanced deep architectures, thereby it might have inferior performance compared to a deep AD model. 

To address the shortcomings and limitations of the prior research works or systems for AI and AD, in this work, we introduce a web-based system for acronym identification and disambiguation that is capable of recognizing and expanding acronyms in multiple domains including general (e.g., Wikipedia articles), scientific (e.g., computer science papers), biomedical (e.g., Medline abstracts), or financial (e.g., financial discussions in Reddit). More specifically, we first propose a rule-based model for acronym identification by extending the set of rules proposed by \cite{schwartz2002simple}. We empirically show that the proposed model outperforms both the previous rule-based model and also the existing state-of-the-art deep learning models for acronym identification on the recent benchmark dataset SciAI \cite{veyseh2020what}. Next, we use a large dataset created from corpora in various domains for acronym disambiguation to train a deep model for this task. Specifically, we employ a sequential deep model to encode the context of the ambiguous acronym and solve the AD task using a feed-forward multi-class classifier. We also evaluate the performance of the proposed acronym disambiguation model on the recent benchmark dataset SciAD \cite{veyseh2020what}.  

To summarize, our contributions are:

\begin{itemize}
    \item The first web-based multi-domain acronym identification and disambiguation system
    \item Extensive evaluation of the proposed model on the two benchmark datasets SciAI and SciAD
\end{itemize}

\section{System Description}

The proposed system is a web-based system consisting of two major components: 1) Acronym Identification which consists of a set of prioritized rules to recognize the mentions of acronyms and their long-forms in the text; 2) Acronym Expansion which involves a dictionary look-up to expand acronyms with only one possible long-form and a pre-trained deep learning model to predict the long-form of an ambiguous acronym using its context. The system takes as input a piece of text and returns the text with highlighted acronyms in which the user can click on the acronyms and their long-form will be shown in a pop-up window. The acronym glossary extracted from the text is also shown at the end of the text. Note that users can also enable/disable the acronym expansion component. This section studies the details of the aforementioned components. 

\subsection{Acronym Identification}

Acronym Identification aims to find the mentions of acronyms and their long-forms in text. This is the first stage in the proposed system to identify the acronyms and their immediate definitions. Generally, this task is modeled as a sequence labeling problem. In our system, we employ a rule-based model to extract acronyms and their meanings from a given text. In particular, the proposed AI model is a collection of rules mainly inspired by the rule introduced in \cite{schwartz2002simple}. More specifically, the following rules are employed in the proposed AI model:

\begin{itemize}
    \item \textbf{Acronym Detector}: This rule identifies all acronyms in text, regardless of having an immediate definition or not. Specifically, all words that at least 60\% of their characters are upper-cased letters and the number of their characters is between 2 and 10 are recognized as an acronym (i.e., short-form).
    \item \textbf{Bounded Schwartz's}: Similar to \cite{schwartz2002simple}, we look for immediate definitions of detected acronyms if they follow one of the templates \textit{long-form (short-form)} or \textit{short-form (long-form)}. In particular, considering the first template, we take the $min(|A|+5,2*|A|)$ words, where $|A|$ is the number of characters in the acronym, that appear immediately before the parentheses as the candidate long-form\footnote{Note that we use the same candidate long-form in other rules too}. Then, a sub-sequence of the candidate long-form that some of its characters could form the acronym is selected as the long-form. However, despite the original Schwartz's rule that does not restrict the first and last word of the long-form to be used in the acronym, we enforce this restriction. This modification could fix erroneous long-form detection by Schwartz's rule. For instance, in the phrase \textit{User-guided Social Media Crawling method (USMC)}, the modified rule identifies the long-form \textit{User-guided Social Media Crawling}, excluding the leading word \textit{method}.
    \item \textbf{Character Match}: While the Bounded Schwartz' rule could identify the majority of the long-forms, it might also introduce some noisy meanings. For instance, in the phrase \textit{Analyzing Avatar Boundary Matching (AABM)}, the Bounded Schwartz's rule identifies \textit{Avatar Boundary Matching} as the long-form of \textit{AABM}, missing the starting word \textit{Analyzing}. To solve this issue and increase the model's accuracy, we also employ a character match rule that assesses if the initials of the words in the candidate long-form could form the acronym. In the given example, it identifies the full phrase \textit{Analyzing Avatar Boundary Matching} as the long-form. Since this rule is more restricted and it has higher precision than Bounded Schwartz's rule, in our system, it has a higher priority than the Bounded Schwartz's rule.
    \item \textbf{Initial Capitals}: One issue with the proposed Character Matching rule is that if there is a word in the long-form that is not used in the acronym, the rule fails to correctly identify the long-form. For instance, in the phrase \textit{Analysis of Avatar Boundary Matching (AABM)} the Character Matching rule fails due to the existence of the word \textit{of}. To mitigate this issue, we propose another high-precision rule, Initial Capitals. In this rule, if the concatenation of the initials of the words of the candidate long-form which are upper-cased could form the acronym, the candidate is selected as the expanded form of the acronym. This rule has the highest priority in our system.
\end{itemize}

In addition to the mentioned general rules, we also add some other rules to handle the special cases (e.g., acronyms with a hyphen, roman numbers, definitions provided in some templates (e..g, CNN stands for convolution neural network)). 

\begin{table}[]
    \centering
    \resizebox{.45\textwidth}{!}{
    \begin{tabular}{l|c|c}
        \textbf{Acronym} & \textbf{Long-form} & \textbf{Rule} \\ \hline \hline
        AABM & Analyzing Avatar Boundary Matching & Character Match \\ \hline
        ABBREX & Abbreviation Expander & Bounded Schwartz's \\ \hline
        AD & acronym disambiguation & Character Match \\ \hline
        AI & Acronym identification & Character Match \\ \hline
        BADREX & Biomedical Abbreviations using & Bounded Schwartz's \\
        & Dynamic Regular Expressions & \\ \hline
        BiLSTM & Bi - directional Long ShortTerm Memory & Bounded Schwartz's \\ \hline
        DOG & Diverse acrOnym Glossary & Bounded Schwartz's \\ \hline
        MAD & Massive Acronym Disambiguation & Capital Initials \\ \hline
        MF & most frequent & Character Match \\ \hline
        USMC & User - guided Social Media Crawling & Capital Initials
    \end{tabular}
    }
    \caption{The acronym glossary extracted from the text of this paper using MadDog.}
    \label{tab:glossary}
\end{table}

\begin{figure}
    \centering
    \includegraphics[scale=0.3]{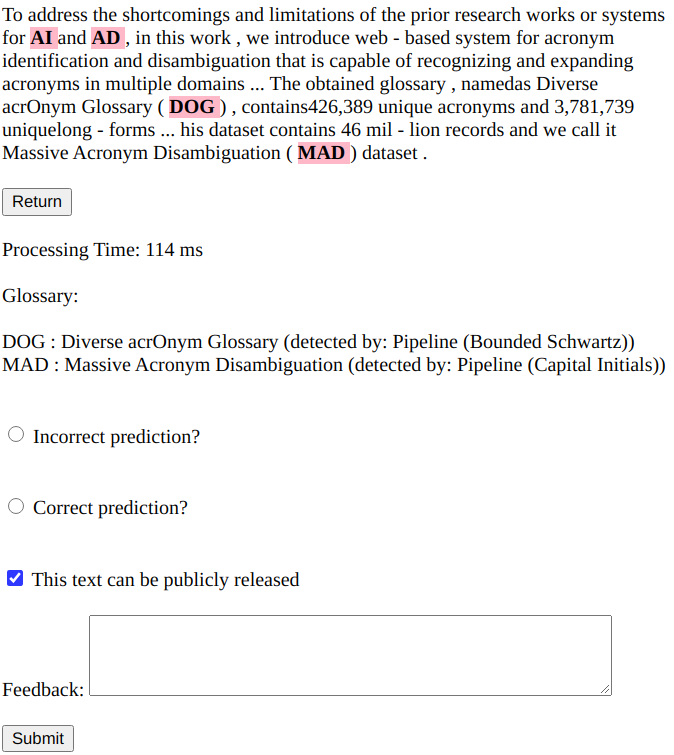}
    \caption{A screenshot of acronym identification by MadDog. It identifies all acronyms and their local-long forms. This interface highlights the detected acronyms and by clicking on them, a pop-up window shows the recognized meaning of the acronym.}
    \label{fig:ai}
\end{figure}

In the web-based system, the user could enter the text and the system recognizes both acronyms without any definition in text and also acronyms that are locally defined with their identified long-forms. Users could also click on each detected acronym to see its definition in a pop-up window. Also, a glossary of detected acronyms and their long-forms is shown at the bottom of the page. A screenshot of the output of the system is shown in Figure \ref{fig:ai}. Moreover, Table \ref{tab:glossary} shows the glossary extracted from the text of this paper using the rule-based component of the system. In section \ref{sec:eval} we compare the performance of the proposed rule-based model with the existing state-of-the-art models for AI \cite{veyseh2020what}.

\subsection{Acronym Expansion}

Although the proposed rule-based model is effective to recognize locally defined acronyms, it might not be able to expand acronyms that don't have any immediate definition in the text itself. To alleviate this issue and expand acronyms even without local definition, two resources are required: 1) A dictionary that provides the list of possible expansion for a given acronym, 2) A model to exploit the context of the given acronym and choose the most likely expansion for a given acronym. For the acronym dictionary, we employ the glossary obtained by exploiting our proposed rule-based AI model on corpora in various domains (i.e., Wikipedia, Arxiv papers,  Reddit submissions, Medline abstracts, and PMC OA subset). The obtained glossary, named as Diverse acrOnym Glossary (DOG), contains 426,389 unique acronyms and 3,781,739 unique long-forms. Note that the previously available web-based acronym disambiguation system \cite{ciosici2018abbreviation} employed only Wikipedia corpus, therefore, it covers limited domains and acronyms compared to our system. 

\begin{table*}[]
    \centering
    \resizebox{.65\textwidth}{!}{
    \begin{tabular}{c|ccc|ccc|c}
        Model & \multicolumn{3}{c}{Acronym} & \multicolumn{3}{c}{Long Form} & \\ \hline
        & P & R & F1 & P & R & F1 & Macro F1 \\ \hline
        NOA & 80.31 & 18.08 & 29.51 & 88.97 & 14.01 & 24.20 & 26.85 \\
        ADE & 79.28 & 86.13 & 82.57 & 98.36 & 57.34 & 72.45 & 79.37 \\
        UAD & 86.11 & 91.48 & 88.72 & 96.51 & 64.38 & 77.24 & 84.09 \\ \hline \hline
        BIOADI & 83.11 & 87.21 & 85.11 & 90.43 & 73.79 & 77.49 & 82.35 \\
        LNCRF & 84.51 & 90.45 & 87.37 & 95.13 & 69.18 & 80.10 & 83.73 \\ 
        LSTM-CRF & 88.58 & 86.93 & 87.75 & 85.33 & 85.38 & 85.36 & 86.55 \\ \hline \hline
        MadDog & 89.98 & 87.56 & 88.75 & 96.45 & 79.53 & 87.18 & 88.12
    \end{tabular}
    }
    \caption{Performance of models for acronym identification (AI)}
    \label{tab:AI}
\end{table*}

In DOG, the average number of long-forms per acronym is 6.9 and 81,372 ambiguous acronyms exist. Due to this ambiguity, a simple dictionary look-up is not sufficient for acronym expansion in the web-based system that uses DOG to expand acronyms with non-local definitions. In order to tackle this problem, we propose to train a supervised model in which the input is the text and the position of the ambiguous acronym in it and the model predicts the correct long-form among all possible candidates. To train this model, we use an automatically labeled dataset obtained by extracting samples from large corpora for each long-form in DOG. This dataset contains 46 million records and we call it the Massive Acronym Disambiguation (MAD) dataset. To split the dataset into train/dev/test splits, we use 80\% of samples of each long-form for training, 10\% for the development set, and 10\% for the test set. It is noteworthy that to facilitate training, before splitting the dataset into train/dev/test splits, we first create chunks of size 100,000 samples in which all samples of an acronym are assigned to the same chunk. Since each acronym appears only in one chuck, we train a separate acronym disambiguation model for each chunk. During inference, we first identify which chuck the ambiguous acronym belongs to, then, we use the corresponding model to predict the expanded form of the acronym.

In this work, we use a deep sequential model to be trained on the MAD dataset for acronym disambiguation. More specifically, given the input text $T=[w_1,w_2,\ldots,w_n]$ with the ambiguous acronym $w_a$, we first represent each word using the corresponding GloVe embedding, i.e., $X=[x_1,x_2,\ldots,x_n]$. Afterward, the vectors $X$ are consumed by a Bi-directional Long Short-Term Memory network (BiLSTM) to encode the sequential order of the words. Next, we take the hidden states of the BiLSTM neurons, i.e., $H=[h_1,h_2,\ldots,h_n]$, and compute the text representation by computing the max-pool of the vectors $H$, i.e., $\bar{h}=MAX\_POOL(h_1,h_2,\ldots,h_n)$. Finally, the concatenation of the text representation, i.e., $\bar{h}$, and the acronym representation, i.e., $h_a$, is fed into a 2-layer feed-forward neural network whose final layer dimension is equal to the total number of long-forms in the dataset (i.e., dataset chunks explained above). 

\begin{figure}
    \centering
    \includegraphics[scale=0.25]{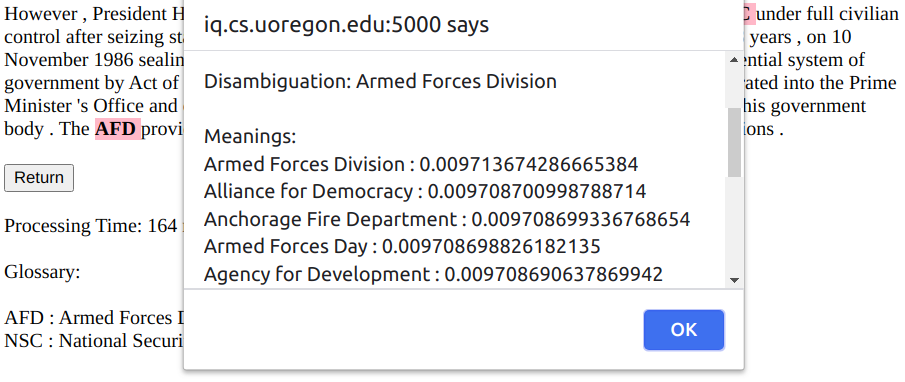}
    \caption{Sorted list of candidate long-forms along with their scores for the acronym \textit{AFD} in the sentence \textit{After 1991, the presidential system of government by Act of Parliament was abolished, and by October 1994, the AFD was integrated into the Prime Minister's Office and concurrently the combined armed forces authority was transferred to this government body.}}
    \label{fig:ad}
\end{figure}

In the proposed system, the long-form of acronyms predicted by the acronym disambiguation model is presented in the glossary at the end of the page (See Figure \ref{fig:ai}). Moreover, by clicking on the acronym word in text, a pop-up window shows the model's prediction and also the sorted list of other candidate long-forms for the selected acronym. An example is shown in Figure \ref{fig:ad}. In the provided example, the system correctly predicts \textit{Gross Domestic Production} as the long-form of the ambiguous acronym \textit{GDP}. We name the proposed acronym identification and disambiguation system as MAdDog.

\section{Evaluation}
\label{sec:eval}
This section provides more insight into the performance of the proposed acronym identification and disambiguation models. To evaluate the performance of the models in comparison with other state-of-the-art AI and AD models, we report the performance of the proposed models on SciAI and SciAD benchmark datasets \cite{veyseh2020what}. We also compare the performance of the proposed model with the baselines provided in the recent work \cite{veyseh2020what}. More specifically, on SciAI, we compare our model with rule-based models NOA \cite{charbonnier2018using}, ADE \cite{li2018guess} and UAD \cite{ciosici2019unsupervised}; and also the feature-based models BIOADI \cite{kuo2009bioadi} and LNCRF \cite{liu2017multi}; and finally the SOTA deep model LSTM-CRF \cite{veyseh2020what}. For evaluation metrics, following prior work, we report precision, recall, and F1 score for the acronym and long-form prediction and also their macro-averaged F1 score. The results are shown in Table \ref{tab:AI}. This table shows that our model outperforms both rule-based and more advanced feature-based or deep learning models. More interestingly, while the proposed model has comparable precision with the existing rule-based models, it enjoys higher recall.  

\begin{table}[]
    \centering
    \resizebox{.35\textwidth}{!}{
    \begin{tabular}{c|ccc}
        Model & P & R & F1 \\ \hline
        MF & 89.03 & 42.2 & 57.26 \\
        ADE & 86.74 & 43.25 & 57.72 \\
        \hline \hline
        NOA & 78.14 & 35.06 & 48.40 \\
        UAD & 89.01 & 70.08 & 78.37 \\
        BEM & 86.75 & 35.94 & 50.82 \\
        DECBAE & 88.67 & 74.32 & 80.86 \\
        GAD & 89.27 & 76.66 & 81.90 \\ \hline \hline
        MadDog & 92.27 & 85.01 & 88.49 
    \end{tabular}
    }
    \caption{Performance of models for acronym disambiguation (AD)}
    \label{tab:AD}
\end{table}

To assess the performance of the proposed acronym disambiguation model, we evaluate its performance on the benchmark dataset SciAD \cite{veyseh2020what} and compare it with the existing state-of-the-art models. Specifically, we compare the model with non-deep learning models including most frequent (MF) meaning \cite{veyseh2020what}, feature-based model (i.e., ADE \cite{li2018guess}), and deep learning models including NOA \cite{charbonnier2018using}, UAD \cite{ciosici2019unsupervised}, BEM \cite{blevins2020moving}, DECBAE \cite{jin2019deep} and GAD \cite{veyseh2020what}. The results are shown in Table \ref{tab:AD}. This table demonstrates the effectiveness of the proposed model compared with the baselines. Our hypothesis for the higher performance of the proposed model is the massive number of training examples for all acronyms which results in low generalization error. 

\section{Related Work}
Acronym identification (AI) and acronym disambiguation (AD) are two well-known tasks with several prior works in the past two decades. For AI, both rule-based models \cite{park2001hybrid,wren2002heuristics,schwartz2002simple,adar2004sarad,nadeau2005supervised,ao2005alice,kirchhoff2016unsupervised} and supervised feature-based or deep learning models \cite{kuo2009bioadi,liu2017multi,veyseh2020what,veyseh2021Acronym} are utilized. Due to the higher accuracy of rule-based models, they are dominantly used in the majority of the related works, especially to automatically create acronym dictionary \cite{ciosici2019unsupervised,li2018guess,charbonnier2018using}. However, the existing works prepare a small-size dictionary in a specific domain. In contrast, in this work, we first improve the existing rules for acronym identification, then, we use a diverse acronym glossary in our system. For acronym disambiguation, prior works employ either feature-based models \cite{wang2016clinical,li2018guess} or deep learning methods \cite{wu2015clinical,antunes2017biomedical,charbonnier2018using,ciosici2019unsupervised,veyseh2021Acronym}. In this work, we also employ a sequential deep learning model for AD. However, unlike prior work that proposes an acronym disambiguation model for a specific domain and limited acronyms, our proposed model covers more acronyms and it is able to expand an acronym in various domains. 

Another common limitation of the existing research-based models for AI and AD is that they do not provide any publicly available system that could be quickly incorporated into a text-processing application. Although there is some software for acronym identification such as expanding Biomedical Abbreviations using Dynamic Regular Expressions (BADREX) \cite{BADREX} or Abbreviation Expander (ABBREX) \cite{ABBREX}, unfortunately, they are incapable of acronym disambiguation. To our knowledge, the most similar work to ours is proposed by  \citet{ciosici2018abbreviation}. Specifically, similar to our work, this web-based system is able to identify and expand acronym in text. A rule-based model is employed for AI and this model is also used to create a dictionary of acronyms. For AD, unlike our work that trains a deep model, they use word embedding similarity to predict the most likely expansion. However, there are some limitations to this previous system. Firstly, it is restricted to the general domain (i.e., Wikipedia) and it covers a limited number of acronyms. Second, it does not provide any analysis and evaluations of the performance of the proposed model. Lastly, it is not publicly available anymore. The proposed MadDog system could be useful for many downstream applications including definition extraction \cite{veyseh2020joint,spala2020semeval,spala2019deft}, information extraction \cite{pouran2019improving,veyseh2020exploiting,veyseh2020multi} or question answering \cite{perez2020unsupervised}

\section{System Deployment}

MadDog is purely written in Python 3 and could be run as a FLASK \cite{grinberg2018flask} server. For text toknization, it employs SpaCy 2 \cite{spacy2}. Also, the trained acronym expansion model requires PyTorch 1.7 and 64 GB of disk space. Note that all acronyms with their long-forms are encoded in the trained model so they can perform both the dictionary look-up operation and the disambiguation task. Moreover, the trained models could be loaded both on GPU and CPU. 

\section{Conclusion}

In this work, we propose a new web-based system for acronym identification and disambiguation. For AI, we employ a refined set of rules which is shown to be more effective than the previous rule-based and deep learning models. Moreover, using a massive acronym disambiguation dataset with more than 46 million records in various domains, we train a supervised model for acronym disambiguation. The experiments on the existing benchmark datasets reveal the efficacy of the proposed AD model.    

\bibliography{anthology,eacl2021}
\bibliographystyle{acl_natbib}

\end{document}